\documentclass[
]{ceurart}

\usepackage{subfig}
\usepackage{listings}
\begin{document}

\copyrightyear{2021}
\copyrightclause{Copyright for this paper by its authors.
  Use permitted under Creative Commons License Attribution 4.0
  International (CC BY 4.0).}

\conference{Forum for Information Retrieval Evaluation, December 13-17, 2021, India}

\title{Contextual Hate Speech Detection in Code Mixed Text using Transformer Based Approaches}

\author[1]{Ravindra Nayak}[%
email=ravindranyk707@gmail.com,
]
\address[1]{Sri Jayachamarajendra College of Engineering, Mysore}

\author[2]{Raviraj Joshi}[%
email=ravirajoshi@gmail.com,
]
\address[2]{Indian Institute of Technology Madras, Chennai}

\begin{abstract}
  In the recent past, social media platforms have helped people in connecting and communicating to a wider audience. But this has also led to a drastic increase in cyberbullying. It is essential to detect and curb hate speech to keep the sanity of social media platforms. Also, code mixed text containing more than one language is frequently used on these platforms. We, therefore, propose automated techniques for hate speech detection in code mixed text from scraped Twitter. We specifically focus on code mixed English-Hindi text and transformer-based approaches. While regular approaches analyze the text independently, we also make use of content text in the form of parent tweets. We try to evaluate the performances of multilingual BERT and Indic-BERT in single-encoder and dual-encoder settings. The first approach is to concatenate the target text and context text using a separator token and get a single representation from the BERT model. The second approach encodes the two texts independently using a dual BERT encoder and the corresponding representations are averaged. We show that the dual-encoder approach using independent representations yields better performance. We also employ simple ensemble methods to further improve the performance. Using these methods we report the best F1 score of 73.07\% on the HASOC 2021 ICHCL code mixed data set. 
\end{abstract}


\maketitle

\section{Introduction}

Social media is a boon to many, as they have helped in creating and promoting budding businesses on such platforms. Although it has vast use cases, it comes with a caveat too. People with malicious intent have considered it as an opportunity to promote hate speech among a wider audience \cite{ezeibe2021hate, matamoros2021racism}. There has been multiple research that directly links social media to poor mental health. Because such platforms are outnumbered by youngsters, their mental stability is trivial in shaping their future careers. So it is necessary to take actions against such malevolent content on a large scale.

Offensive language such as insulting, hurtful, derogatory or obscene content directed towards people might suppress meaningful discussions. As there are no restrictions on expressing peoples opinions on such platforms, it might lead to the defaming of personalities. So it is the platform’s responsibility to restrain such content. Hate speech mainly involves discriminating against people based on religion, community, race, nationality, gender or any other identity factors \cite{info12010005, macavaney2019hate}.

Even though manual moderation of hate speech is always precise, it isn't recommended considering the huge volumes of data that is being pumped into social media. So there is a constant need for automated techniques to suppress such hateful content where the ages of all the groups are exposed to \cite{schmidt2017survey,hasoc2021mergeoverview,joshi2021evaluation,wani2021evaluating}. As we have seen advances in computing capabilities, machine learning algorithms have gained their importance in tasks that involve understanding natural language.

In this work, we are interested in the hate speech detection of tweets. This paper mainly focuses on evaluating the HASOC 2021, Identification of Conversational Hate-Speech in Code-Mixed Languages (ICHCL) subtask \cite{hasoc2021ICHCLoverview}. This task aims to detect hate speech in individual tweets and their respective comments and replies which support hate speech directly or indirectly. The dataset contains scraped text from Twitter with binary labels. A conversational thread can contain abusive or offensive content, which is not apparent just from a single comment or the reply to a comment but can be identified if given the context of the parent content as shown in Table \ref{examples}. Furthermore, the contents on such social media are spread in so many different languages, including code-mixed languages such as Hinglish (mix of Hindi and English in roman text) \cite{joshi2020evaluating}.

\begin{table}
\caption{Few training samples of context and tweets. HOF indicates Hate \& Offensive speech, whereas NOT indicates Non-Hate Offensive speech content. }
\label{examples}
\begin{tabular}{p{5cm}p{7.5cm}p{1cm}}
\hline
    Context & Tweet & Label \\
\hline
    - & INDIA NEEDS VACCINES & NOT \\
    INDIA NEEDS VACCINES & Yes ma'am India need vaccine Doctors and facilities & NOT \\
    INDIA NEEDS VACCINES & Is there any Vaccine which can prevent India from you & HOF \\
    INDIA NEEDS VACCINES & vaccine insano k liye hain reptiles k liye nhi & HOF \\
\hline
    - & Look at this insensitive piece of shit & HOF \\
    Look at this insensitive piece of shit & After all that is happening since last month but still & NOT \\
    Look at this insensitive piece of shit & Yo wtf & HOF \\
    Look at this insensitive piece of shit & Haha & HOF \\
\hline
\end{tabular}
\end{table}

The hate speech detection task can be considered as a binary text classification problem \cite{joshi2019deep}. We solely depend upon the tweets and the context of the tweet to determine hateful content. Even though some of the tweets can be rejected considering the behaviour of the content creator, we cannot always guarantee that this information is available every time. We evaluate various deep learning techniques, specifically the multilingual BERT based models. We have tried to experiment on various fine-tuning methods, and how they are helpful for the model to detect malicious tweets.

\section{Related Work}

Hate speech detection is precise when manually moderated. The context of such tweets is also important to identify hatefulness. For the code mixed data, in particular, the moderator must have a vast knowledge of vocabulary across languages to curb malicious content. If we have enough data on the user's behaviour and tweet content, then it could help us in mitigating such content by blacklisting such users. Many approaches like graph convolution networks are being used that capture not only the structure of online communities but also the linguistic behaviour of the users within them \cite{mishra2019abusive}.

Dictionary-based approaches are popular for text data where we try to maintain a list of words or phrases that might be profane or any kind of racial slurs. Various machine learning approaches involve the usage of extra-linguistic features in conjunction with character n-grams to build binary logistic regression classifiers \cite{waseem-hovy-2016-hateful}. There have been studies showing that including knowledge graph features have helped in building better models \cite{10.1145/3430984.3431072}.

Word level embeddings like Glove have helped in better capture of the semantics of words in comparison to one-hot encoding \cite{zhang2018hate}. Another similar approach is to make use of sentence-level embeddings like ELMo which help in extracting rich features from the text. These embeddings are then fed to bi-directional LSTMs or CNNs for classification \cite{rizoiu2019transfer}. As these embeddings are trained on huge corpora of data, they are often called transfer learning as they help in reusing feature-rich vectors for similar classification tasks. Various other features like LIWC features, SentiWordnet and Profanity vectors also aid the model \cite{mathur-etal-2018-offend}.

For the code mixed Hinglish data sets, there have been studies on ensembling BERT based embeddings along with Bi-LSTM to improve the model \cite{info12010005}. As context plays an important role in the detection of hate speech, context-aware models are built which take previous tweet’s features as an input along with the current tweet. Various ensembles of traditional machine learning algorithms with deep learning techniques have also been explored \cite{gao-huang-2017-detecting}.

\section{Architecture details}

In this section, we describe the details of different techniques along with their hyperparameters. Figure \ref{fig1} gives a summary of the model details along with 2 architectures that were explored in this work.

We use transformer-based neural networks as they have shown great progress in NLP tasks \cite{vaswani2017attention}. As these networks help in the parallelisation of computations, they have an immense advantage over their predecessor networks like RNN and  LSTM. Transformers reduce the latency of model inference time as they are capable of making use of the contemporary hardware available. We explore two multilingual variations of BERT-based models viz. m-BERT and Indic-BERT. Both the BERT variations include Hindi as one of the pre-training languages.

\subsection{Multilingual-BERT (m-BERT)}
This model’s architecture is based on BERT-base \cite{devlin2019bert}. It is a model that contains 12 transformer blocks, 12 self-attention heads, hidden size of 768. The input for BERT contains a maximum embedding of 512 words and it outputs a sequential representation. Special tokens like [CLS] and [SEP] are used to specify the start of a sentence and separation of sentences respectively. For a classification task, final encoder representations are considered and a  softmax is applied to classify the representation. 

As the BERT-base consists of only English text, we use a Multilingual BERT-base model that has been trained on 102 languages using a shared word-piece vocabulary of size 110k. Oversampling of low resource languages is done to overcome data imbalance. It has shown great results on zero-shot transfer learning for various downstream tasks and also helped in code-switched data tasks \cite{pires2019multilingual}.

\subsection{Indic-BERT}

This model is based on ALBERT \cite{lan2020albert}, which is a lighter version of BERT as they incorporate parameter sharing across layers which in turn leads to lesser parameters. They have also made modifications in pre-training mechanisms by introducing new pre-training tasks that have led to better sentence embeddings. ALBERT contains 12 transformer blocks, 12 self-attention heads, a hidden size of 768 and an input embedding size of 128.

Indic-BERT is a multilingual ALBERT based model that has been trained on 12 major Indian languages with a shared vocabulary size of 200k. It has outperformed multilingual BERT in some of the Indic tasks \cite{kakwani2020indicnlpsuite}.

\begin{figure}
  \centering
  \includegraphics[width=\linewidth]{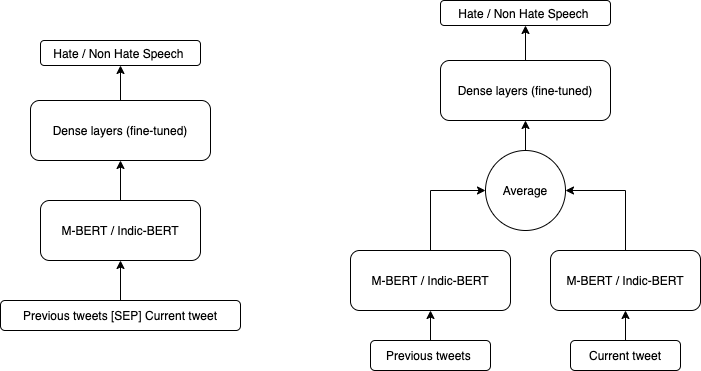}
  \caption{Representation of single encoder approach (left) and dual encoder approach (right).}
  \label{fig1}
\end{figure}

\section{Experimental Setup}

\subsection{Dataset details}
The HASOC 2021 ICHCL dataset \cite{hasoc2021ICHCLoverview} consists of tweets and their context if any, along with the labels. The binary labels consist of (NOT) Non-Hate Offensive and (HOF) Hate and Offensive. This dataset comprises 2 level hierarchy where an individual tweet can be followed by a comment, and that comment can have a reply. In the case of comments, we consider individual tweets as the context, and for replies, we consider the concatenated context of the first tweet and the comment associated with it.

The dataset mainly consists of a train and test set. There are a total of 7088 tweets provided as a dataset, out of which there are 5740 training samples and 1348 test samples. We have considered a random 10 per cent of the data for validation purposes. Training data contains 2841 hate speech samples and 2899 non-hate speech samples, whereas test data contained 695 hate speech samples along with 653 non-hate speech samples. As the task mainly focuses on context-based hate speech detection, there were 82 individual tweets in training and 16 individual tweets in testing. The remaining 5658 data points in training and 1332 data points in the test used the individual tweets as the context. More statistics on the data is provided in Table \ref{stats}.

\begin{table}
\caption{Statistics of the dataset}
\label{stats}
\begin{tabular}{c c c}
\hline
    Feature & Train & Test  \\
\hline
    Total words & 273627 & 68019 \\
    Max word length & 166 & 131 \\
    Avg word count & 47.67 & 50.45 \\
    Unique tokens & 5717 & 1343 \\
\hline
\end{tabular}
\end{table}

\subsection{Data preprocessing}
Various data preprocessing techniques are used to clean and normalize the tweets.
\begin{itemize}
    \item \textbf{Removal of URLs}: Often people use hyperlinks to different websites. As this might not help us, we are removing it.
    \item \textbf{Removal of User Mentions}: User mentions are commonly used in tweets. Their removal is necessary as it is not helpful to the model.
    \item \textbf{Removal of Non-Hindi and Non-English characters}: As we are sure about the dataset containing only Roman and Devanagari text, we remove characters outside the Unicode block.
    \item \textbf{Retain Emojis and Hashtags}: We retain emojis and hashtags, as this will help in determining whether a tweet is supporting a hateful tweet, in the absence of text.
\end{itemize}

\subsection{Training details}
All the models were trained using the PyTorch framework and Hugging Face library \cite{wolf2020transformers}. The models have been finetuned up to a maximum of 5 epochs and the minimum validation loss is the criteria used for picking the best epoch.
As discussed in Figure \ref{fig1}, we mainly work on 2 approaches for m-BERT and Indic-BERT.

\begin{itemize}
\item \textbf{Single Encoder Approach (single sentence representation)}: This is a basic approach of fine-tuning BERT based models, where we add a dense layer after the BERT [CLS] token embedding followed by the softmax classifier. The context text and the target text are concatenated using a separator token to get a single [CLS] representation from the BERT model.
\item \textbf{Dual Encoder Approach (averaging the context and target representations)}: As context plays a vital role in our dataset, we passed the context and the tweet separately to the BERT to get their [CLS] token embeddings. This embedding acts as a sentence representation for individual context and tweets. These embeddings are averaged and further passed to the dense layer for classification. If the context is absent, then we consider only the tweet representations.

\end{itemize}

\begin{table}
\caption{Result metrics for different BERT configurations. FE indicates BERT model with frozen embedding layer. C-Avg indicates averaging over the [CLS] token embeddings of the context and tweet. Dictionary indicates a static list of profane words.}
\label{results}
\begin{tabular}{c c c c}
\hline
    Model & Precision & Recall & F1 score  \\
\hline
    m-BERT baseline & 66.07 & 65.63 & 65.53 \\
    Indic-BERT baseline & 67.18 & 67.17 & 67.17 \\
\hline
    m-BERT + frozen embeddings (FE) & 70.03 & 67.40 & 66.65 \\
    m-BERT + FE + C-Avg & 67.70 & 67.65 & 67.65 \\
    m-BERT + FE + C-Avg + Dictionary & 68.82 & 68.62 & \textbf{68.61} \\
\hline
    Indic-BERT + FE + Dictionary & 70.71 & 70.07 & 69.99 \\
    Indic-BERT + FE + C-Avg + Dictionary & 71.09 & 70.44 & \textbf{70.37} \\
\hline
    Ensemble 2 (Indic-BERT C-Avg+ m-BERT C-Avg) & 71.65 & 71.59 & 71.60 \\
    Ensemble 4 (Indic-BERT + Indic C-Avg + m-BERT + m-BERT C-Avg) & 73.21 & 73.17 & \textbf{73.07} \\
\hline
\end{tabular}
\end{table}

\section{Results and Discussions}
\begin{figure}
  \centering
    \subfloat{\includegraphics[scale=0.5]{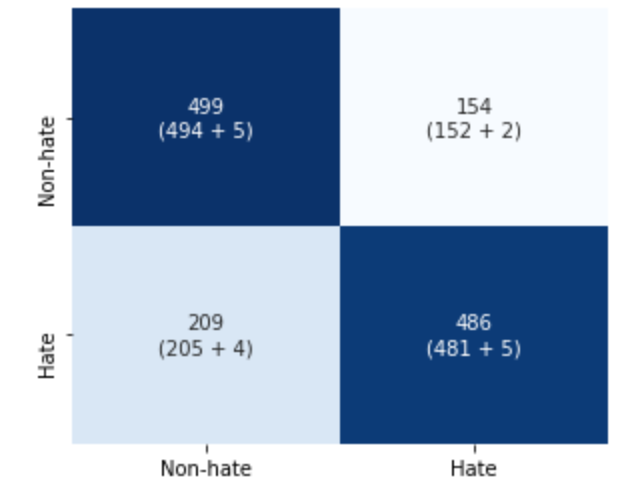}}
    \hspace{2cm}
    \subfloat{\includegraphics[scale=0.5]{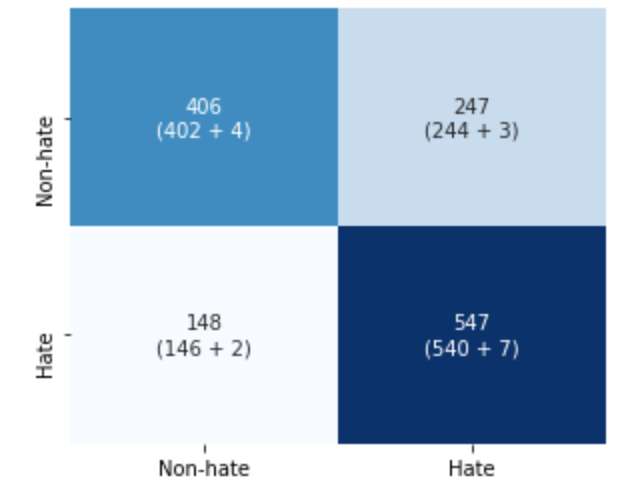}}
  \caption{Confusion matrices for the Ensenble 4 approach (left) and IndicBERT Dual Encoder approach (right). The rows correspond to true class and columns correspond to predicted class. Each cell value further is segregated as the number of contextual examples + the number of non-contextual examples.}
  \label{fig:confusion}
\end{figure}

We evaluate different BERT-based approaches for the task of Hate speech detection. The results of the experiments are outlined in Table \ref{results}. The macro precision, recall, and F1-scores are metrics used to compare the models. As the target text uses code-mixed Hindi and English language, we use m-BERT and Indic-BERT as our baselines. In the baseline approach, we concatenate the target and the context text using a separator token. We perform a series of experiments on top of the baseline model by freezing the embedding layer and incorporating a static dictionary of offensive words. The frozen embeddings showed promising results as the token embeddings were not overfitted to the training data. The static dictionary is used as a deterministic classifier by directly tagging a text as hateful if any offensive word is present in the text. The dictionary was created using various web sources and neither train data nor test data were referenced during the process. In the dual encoder approach, we average out the [CLS] token embeddings for the context and the tweet, further showing improvement in F1 scores. Integration of the static dictionary with this method further improves the F1 numbers. The Indic-BERT model with frozen embeddings, static dictionary, and dual representations approach outperformed all the other techniques. We combine the best-performing models using simple ensemble techniques to get the best results. The scores of the individual models are fused using averaging. The confusion matrices for best models are shown in Figure \ref{fig:confusion}

\section{Conclusion}
Under the HASOC 2021 ICHCL task, we try to evaluate various finetuning techniques for code mixed data considering the context of the tweets. We have mainly focused on multilingual BERT based architectures. We observe that frozen embeddings give better results by retaining rich token representations from the pre-trained model. Moreover, averaging over sentence representations has helped the model in understanding the context better while trying to classify the current tweet. Using the ensemble of models, we have achieved the best F1 score of 73.07\%, over the m-BERT baseline F1 score of 65.53\%. Primarily we emphasize the importance of averaging representations using the dual BERT encoder setting in context-based text classification problems.

\begin{acknowledgments}
This research was conducted under the guidance of L3Cube, Pune. We would
like to express our gratitude towards our mentors at L3Cube for their continuous support and encouragement. 
\end{acknowledgments}

\bibliography{hasoc}

\appendix

\end{document}